\documentclass[letterpaper, 10 pt, conference]{ieeeconf}

\IEEEoverridecommandlockouts                             
\usepackage[pdftex]{graphicx}

\graphicspath{{figure/}}
\usepackage{amsmath}
\usepackage{amssymb}
\usepackage{url}

\usepackage{amsthm} 
\usepackage{cite}
\usepackage{color}
\usepackage{comment}
\usepackage{fancyhdr} 

\usepackage{algorithm}
\usepackage{algorithmic}
\usepackage{bm}
\usepackage{bbm}
\let\labelindent\relax
\usepackage{enumitem}
\usepackage{mathtools}

\newtheorem{theorem}{\textbf{Theorem}}
\newtheorem{lemma}{\textbf{Lemma}}
\newtheorem{assumption}{\textbf{Assumption}}

\newcommand{\sA}{\mathcal{A}}

\newcommand{\sG}{\mathcal{G}}

\newcommand{\sS}{\mathcal{S}}      
\newcommand{\sT}{\mathcal{T}}

\newcommand{\R}{\mathbb{R}}                     

\newcommand{\ba}{\bm{a}}

\newcommand{\bs}{\bm{s}}

\newcommand{\bla}{\bm{\lambda}}

\newcommand{\pri}{\mathrm{Pri}}
\newcommand{\rel}{\mathrm{Rel}}
\newcommand{\ind}{\mathrm{Ind}}

\allowdisplaybreaks

\title{\bf Contextual Restless Multi-Armed Bandits  with Application to \\   Demand Response Decision-Making}

\author{Xin Chen, I-Hong Hou
\thanks{ X. Chen and I. Hou are with the Department of Electrical and Computer Engineering, Texas A\&M University, TX, USA. Emails: {\tt\small xin\_chen@tamu.edu}, {\tt\small ihou@tamu.edu}. }
\thanks{ 
The work was supported in part by NSF under Award Number ECCS-2127721, in part by the U.S. Army Research Laboratory and the U.S. Army Research Office under Grant Number W911NF-22-1-0151, and in part by Office of Naval Research under Contract N00014-21-1-2385..} 
}

\begin{document}

\maketitle

\begin{abstract}
This paper introduces a novel multi-armed bandits framework, termed Contextual Restless Bandits (CRB), for complex online decision-making. This CRB framework incorporates the core features of contextual bandits and restless bandits, so that it can model both the internal state transitions of each arm and the influence of external global environmental contexts. Using the dual decomposition method, we develop a scalable index policy algorithm for solving the CRB problem, and theoretically analyze the asymptotical optimality of this algorithm. In the case when the arm models are unknown, we further propose a model-based online learning algorithm based on the index policy to learn the arm models and make decisions simultaneously. Furthermore, we apply the proposed CRB framework and the index policy algorithm specifically to the demand response decision-making problem in smart grids. The numerical simulations demonstrate the performance and efficiency of our proposed CRB approaches.

\end{abstract}




\section{Introduction}

Multi-armed bandits \cite{slivkins2019introduction} is a prominent online decision-making framework for algorithms that make sequential decisions under uncertainty. It
has found extensive applications across a broad spectrum of domains \cite{bouneffouf2019survey}, such as dynamic pricing \cite{misra2019dynamic}, demand response \cite{chen2020online,li2020reliability},  clinical trials \cite{villar2015multi}, and recommendation systems \cite{elena2021survey}. Moreover, various extensions of multi-armed bandits have been developed to further enhance its modeling capabilities. In particular, \emph{restless}  bandits \cite{whittle1988restless,weber1990index} extends the framework by incorporating the internal state and dynamic state transitions of each arm, treating each arm as a Markov decision process (MDP). 
\emph{Contextual} bandits \cite{slivkins2019introduction,zhou2015survey} leverages observed side information, i.e., contexts, and model their impacts on the rewards of arms, with the goal of learning 
optimal policies that map contexts to arms.
Despite extensive research and numerous successful implementations, the approaches of restless bandits and contextual bandits have been studied and applied separately.

However, in many practical applications, each agent (arm) not only has its own internal state and transition dynamics but is also influenced by external environmental contexts. In this paper, we focus on the demand response (DR) application \cite{chen2020online} in smart grids \cite{chen2022reinforcement},
using it as a case example to illustrate the underlying motivation and our solution development. To enhance reliable electricity supply,
DR \cite{qdr2006benefits} serves as an effective mechanism that calls for the adjustment of flexible load demands from end-users to fit the needs of power grids. In an incentive-based DR program,  the load service entity (LSE) anticipates an upcoming DR event, typically the time of a load peak or power grid emergency, 
and then signals users to reduce their load with certain incentives, e.g., \$5. Since each user's DR participation comes with a cost, it is crucial for the LSE to select the right users to signal for maximizing the total load reduction under a financial budget. Hence, the DR decision-making problem (i.e., optimal user selection) can be aptly modeled using the multi-armed bandits framework \cite{chen2020online,li2020reliability}, where each user is treated as an arm and the LSE takes arm selection actions.


A critical challenge in the DR decision-making process is that users' load reduction behaviors are uncertain and unknown to the LSE. A number of
studies \cite{xu2018promoting,fell2015public,shi2019estimating} suggest that 
users' DR behaviors exhibit significant individual diversity and are influenced by various external environmental contexts, including ambient temperature, weather conditions, electricity prices,  offered incentives, the time of day, etc. Furthermore,
users' DR behaviors are not static but evolve over time and are affected by the selection actions of the LSE.  
 For example,  it is widely observed in real-world DR programs that users tend to become less responsive to the LSE's signals after consecutive participation in DR events, a phenomenon known as \emph{user response fatigue} \cite{kim2011common,li2020reliability}. In addition, flexible load devices 
 are characterized by specific physical states and control dynamics, e.g., the on/off states and switching dynamics of thermostatically controlled loads \cite{taylor2013index,chen2021online}, and the battery state of charge (SOC) and charging dynamics of electric vehicles \cite{yu2018deadline}. Hence, it is critical to incorporate models that account for both the influence of external environmental contexts and the 
internal states and transitions of users for optimal DR decision-making.

 Regarding the DR application, references \cite{taylor2013index,yu2018deadline,yu2016demand} formulate the flexible load control and scheduling in DR programs 
 as restless bandits problems and derive index policies for optimal decision-making. In our previous work \cite{chen2020online}, the optimal user selection is modeled 
in the contextual bandits framework to capture the influence of
time-varying environmental contexts. However, these studies exclusively model either the internal state transitions or the external global contexts, but not both aspects concurrently. 
In the multi-armed bandits literature, there is very limited existing work addressing the restless bandits with the integration of global contextual information. A recent work \cite{liang2024bayesian} enhances the Bayesian learning efficiency in the restless bandits problem by incorporating the contextual information shared within and across arms, while it differs from this paper that focuses on the influence of global environmental contexts.

\vspace{1pt}
 \emph{Our Contributions}. Motivated by this critical issue, we
 develop a novel multi-armed bandits framework, termed \emph{Contextual Restless Bandits} (CRB), for complex online decision-making, by integrating the core features of contextual bandits and restless bandits. In the proposed CRB framework, we represent each arm as a context-augmented MDP to describe its internal states and transitions; additionally, we introduce a global context, modeled as a Markov process, to encapsulate the influence of external environmental factors on arms. The key contributions of this paper are threefold: 
 {\setlist{leftmargin=4.5mm}
\begin{itemize}
    \item [1)] To our knowledge, this is the first work that introduces a multi-armed bandits framework that incorporates both the internal arm state transitions and the influence of global contexts, with significantly enhanced modeling capability.
    \item [2)] We develop a scalable index policy algorithm (see Algorithm \ref{alg:index_policy}) for solving the CRB problem and theoretically analyze the asymptotic optimality of the algorithm. Built upon it, an online learning algorithm (see Algorithm \ref{alg:learn}) is further proposed to solve the CRB problem in the case when the models of arms are unknown.
    
    \item [3)] We tailor the proposed CRB framework and index policy algorithm specifically for the DR decision-making problem, and demonstrate the algorithm's efficiency and performance via numerical simulations.

\end{itemize}
 }

The remainder of this paper is structured as follows. Section \ref{sec:problem} introduces the CRB framework and its application to DR decision-making. Section \ref{sec:algorithm} develops the index policy and online learning algorithms. Section \ref{sec:analysis} analyzes the asymptotic optimality. Numerical experiments are conducted in Section \ref{sec:simulation}, and Section \ref{sec:conclusion} concludes this work.

\vspace{2pt}
\emph{Notations}. Denote $[N]\!:=\!\{1,2,\cdots,N\}$ for an integer $N$. Let $\mathbb{P}[\cdot]$ and $\mathbb{E}[\cdot]$  denote  the probability and expectation.

\section{Problem Formulation} \label{sec:problem}

This section first introduces the CRB framework model and then adapts it to the DR 
decision-making problem.

\subsection{Contextual Restless Bandits (CRB)}

In the CRB framework, a decision-maker interacts with $N$ arms in an infinite time horizon $\sT:=\{0,1,\cdots\}$. 
 Each arm $i\in[N]$ is modeled as a context-augmented
MDP, defined by the state space 
$\sS$, action space $\sA$, global context space $\sG$, reward function $R_i(g,s,a):\sG\times\sS\times\sA\to \R$, and the transition probability function $P_i(s'|g,s,a)\!:=\!\mathbb{P}(s_{i,t+1}\!=\!s'|g_{t}\!=\!g,s_{i,t}\!=\!s,a_{i,t}\!=\!a)$. We consider that $\sS,\sA,\sG$ are \emph{discrete} and shared across all arms, and the action space is \emph{binary} with $\sA\!:=\!\{0,1\}$, where $a\!=\!1$ means that the arm is selected by the decision-maker to activate and otherwise $a\!=\!0$. 
Moreover, we assume that the evolution of the global context $g\in\sG$ follows a positive-recurrent Markov process that is independent of the states and actions of arms, which is defined by the transition probability function $G(g'|g)\!:=\! \mathbb{P}(g_{t+1}\!=\!g'|g_t\!=\!g)$. 

At each time $t\in\sT$, the decision-maker observes the
state $s_{i,t}\in\sS$ of each arm $i\in[N]$ and 
the global context $g_t\in\sG$. Then, the decision-maker selects at most $C_{g_t}$ arms to activate, i.e.,  $\sum_{i=1}^N a_{i,t} \leq C_{g_t}$, where we assume that the budget $C_{g_t}$ depends on the global context $g_t$. 
After that, regardless of being selected or not, each arm $i$ generates a reward $r_{i,t}\!=\!R_i(g_t,s_{i,t},a_{i,t})$ and transitions to the next state $s_{i,t+1}$ according to the probability function $P_i(s'|g,s,a)$, while
the global context evolves to $g_{t+1}$ according to $G(g'|g)$. Denote $\bs\!:=\!(s_1,\cdots,s_N)$ and $\ba\!:=\!(a_1,\cdots,a_N)$ as the 
tuple of states and actions of all arms. 
The goal of the decision-maker is to design a control policy $\pi(\ba|\bs,g)$ to maximize the long-term cumulative discounted reward. This goal is formulated as the following Primal problem \eqref{eq:primal}: 
\begin{subequations} \label{eq:primal}
    \begin{align}
(\textbf{Primal}):  \ \ &
\max_{\pi}\,   \mathbb{E}_\pi\Big[\sum_{t=0}^\infty \sum_{i=1}^N \beta^t r_{i,t}\Big] \label{eq:primal:obj}\\
    \text{s.t. } & \sum_{i=1}^N a_{i,t}\leq C_{g_t},   &&  \hspace{-35pt} \forall t\in\sT, \label{eq:primal:budget}
\end{align}
\end{subequations}
where $\beta\in(0,1)$ is the discount factor and $ a_{i,t}\in\{0,1\}$.
Hence, the selection decisions made by the decision-maker are affected by both the global context (contextual) and the individual state transitions of each arm (restless).

\emph{Model Availability.}
We assume that the transition kernel $G$ of the global context is \emph{known} or can be accurately estimated offline, as ample historical observations of the global context are typically available in practice. 
The reward function $R_i$ is assumed to be known as well. 
In terms of the transition kernel $P_i$ of each arm $i$, we consider two problem settings: 
\begin{itemize}
    \item [1)] (\emph{Known Model}): $P_i$ is known for all $i\in[N]$;
    \item [2)] (\emph{Unknown Model}):  $P_i$ is unknown for all $i\in[N]$.
\end{itemize}

As presented in Section \ref{sec:algorithm}, when the model is known, we can directly derive the optimal control policy (see Algorithm \ref{alg:index_policy}) through dynamic programming and dual decomposition; when the model is unknown, we propose an online CRB learning algorithm (see algorithm \ref{alg:learn}) to learn the model of $P_i$ while concurrently making arm selection decisions.

\subsection{Application of CRB to Demand Response}
  \label{sec:drapp}

Consider the  DR program 
where an aggregator (decision-maker) coordinates $N$ users (arms) for load reduction. Each time $t\in\sT$ corresponds to a DR event \cite{chen2020online}, during which the aggregator selects a set of users to signal for load reduction. 

 {\setlist{leftmargin=3mm}
\begin{itemize} 
 
    \item \emph{Action}: $a_{i,t}\in \sA:=\{0,1\}$ denotes the action whether the aggregator selects user $i$ to signal or not at time $t$.
    
    \item \emph{State}: Each user $i$ has the individual state $s_{i,t}$ defined as:
    \begin{align}\label{eq:state}
     \quad   s_{i,t}:=( z_{i,t}, x_{i,t}), \quad \forall i\in[N], t\in\sT,
    \end{align}
    where $z_{i,t} \!\in\! \{0,1\}$ is a binary state indicating whether user $i$ would reduce their load or not once being signaled. $x_{i,t}$ denotes other user states that are affected by the actions, such as the user fatigue level. For example, $x_{i,t}$ can include the number of user $i$'s DR participation in the most recent $K$ DR events to account for the fatigue effect. Additionally, the physical system dynamics of load devices can also be captured by $x_{i,t}$, e.g., the switching dynamics of thermostatically controlled loads \cite{taylor2013index,chen2021online}.

\item \emph{Global Context}: $g_t$ denotes the exogenous environmental factors at time $t$ that influence users' state transitions and are commonly shared among users, such as temperature, weather conditions, electricity price, DR incentive credits, time of day, day of week, grid conditions, and so forth.

\item \emph{Reward}: Let $\ell_i(g_t,s_{i,t})$ be the load reduction that user $i$ is willing to undertake, which is a function of the global context and internal state. For example, users typically reduce less load for cooling when the ambient temperature is higher in summer. Then, the reward function of user $i$
is given by $R_i(g_t,s_{i,t},a_{i,t})\! =\! a_{i,t} z_{i,t}\ell_i(g_t,s_{i,t})$, which is the actual reduced load from user $i$ if being signaled.

 \item \emph{Goal}: In practice, an aggregator may coordinate thousands to millions of users with flexible loads. It aims to select a subset of users to send load curtailment signals in order to maximize the total load reduction. Since each user's participation entails a cost \cite{chen2021online}, such as an incentive credit of \$5, the aggregator operates with a financial budget that limits the number of selected users, e.g., to $C_{g_t} \!=\! 200$.


\end{itemize}
}

In this way, the DR problem is effectively modeled using the CRB framework, capturing both the impact of global environmental context and internal state transitions of individual users. Section V presents the numerical simulations conducted for the CRB-based DR decision-making.

\section{Index Policy Design via Dual Decomposition} \label{sec:algorithm}

In this section, we introduce an index policy to solve the Primal problem \eqref{eq:primal}, when the dual decomposition method is employed to address the issue of scalability. 


\subsection{Dual Decomposition of Primal Problem}

The Primal problem \eqref{eq:primal} involves a high-dimension MDP whose state space is $\sS^N\times \sG$ and expands exponentially as the number of arms $N$ increases. Hence, we employ the dual decomposition method \cite{adelman2008relaxations} to 
 tackle the challenge of high dimensionality. We first relax the per-time-step budget constraint \eqref{eq:primal:budget} into an expected per-global-context budget constraint \eqref{eq:relax:budget}.  Specifically, constraint \eqref{eq:relax:budget} indicates that for every possible global context $g\in\sG$, the number of selected arms \emph{on average} does not exceed $C_g$, while the actual number of selected arms could be larger than $C_g$ at certain times $t$.  
 As a result, the Primal problem \eqref{eq:primal} is relaxed as the Relaxed problem \eqref{eq:relax}:
 \begin{subequations} \label{eq:relax}
    \begin{align}
(\textbf{Relaxed}): \ \  &
\max_{\pi}\,   \mathbb{E}_\pi\Big[\sum_{t=0}^\infty \sum_{i=1}^N \beta^t r_{i,t}\Big] \label{eq:relax:obj}\\
    \text{s.t. } &  \mathbb{E}_\pi\Big[\sum_{t=0}^\infty \beta^t\mathbb{I}(g_t=g)\big(\sum_{i=1}^N a_{i,t}\big)\Big]\nonumber\\
    &\leq \mathbb{E}_\pi\Big[\sum_{t=0}^\infty \beta^t\mathbb{I}(g_t=g)\Big]C_g,\  \  \forall g\in\sG,  \label{eq:relax:budget}
\end{align}
\end{subequations}
 where $\mathbb{I}(\cdot)$ is the indicator function.
 
 To solve the Relaxed problem \eqref{eq:relax}, we introduce a Lagrange multiplier $\lambda_g\!\geq\! 0$ for each relaxed constraint \eqref{eq:relax:budget} under the global context $g\in\sG$, and obtain the following Lagrange function $L(\pi,\bla) $ \eqref{eq:lagrang} with $\bla\!:=\!(\lambda_g)_{g\in\sG}$.
     \begin{align}  \label{eq:lagrang}
&L(\pi,\bla) =  \mathbb{E}_\pi\Big[\sum_{t=0}^\infty \sum_{i=1}^N \beta^t r_{i,t}  \nonumber \\
    &\qquad -\sum_{g\in\sG}\lambda_g\Big(\sum_{t=0}^\infty \beta^t\mathbb{I}(g_t=g)\big(\sum_{i=1}^Na_{i,t}-C_g\big)\Big)\Big]\nonumber\\
&=\!\sum_{i=1}^N\!  \mathbb{E}_\pi\Big[\sum_{t=0}^\infty\beta^t(r_{i,t}\!-\!\lambda_{g_t}a_{i,t}) \Big]\! +\! \mathbb{E}_\pi\Big[\sum_{t=0}^\infty\!\beta^t\lambda_{g_t}C_{g_t}\Big].
\end{align}
In this way, the Relaxed problem \eqref{eq:relax} is converted to the Dual problem \eqref{eq:saddle}: 
\begin{align}\label{eq:saddle}
(\textbf{Dual}):  \ \  \min_{\bla\geq \bm{0}}  \Big[ \max_{\pi}  \, L(\pi,\bla) \Big].
\end{align}
Reference \cite{adelman2008relaxations} has shown that strong duality holds when the state space and the action space are both finite.
Given a Lagrange multiplier $ {\bla}$, the corresponding optimal policy $\pi^*( {\bla}) $ can be obtained by solving $  \max_{\pi} L(\pi, {\bla})$, i.e., the inner optimization of \eqref{eq:saddle}. In this case, since the global context $g_t$ is independent from the decisions made by the control policy $\pi$, 
the second term $\mathbb{E}_\pi\Big[\sum_{t=0}^\infty\beta^t\lambda_{g_t}C_{g_t}\Big]$
in \eqref{eq:lagrang} is a constant for optimizing $\pi$ and thus can be removed. Consequently, solving $\max_{\pi} L(\pi, {\bla})$ can be equivalently decomposed into $N$ independent sub-problems, each corresponding to an arm.  
Specifically, the sub-problem associated with arm $i\in[N]$ is formulated as \eqref{eq:subprob}:
\begin{align} \label{eq:subprob}
   (\textbf{Arm}_i( {\bla})): \ \max_{\rho_i} \  &\mathbb{E}_{\rho_i}\Big[\sum_{t=0}^\infty\beta^t(r_{i,t}- {\lambda}_{g_t}a_{i,t})\Big], 
\end{align}
where $\rho_i(a|g,s,  {\bla})$ denotes the policy for arm $i$ that selects the action $a_{i,t}\in\{0,1\}$ based on $(g_{t},s_{i,t})$ and the given Lagrange multiplier $ {\bla}$. Solving each sub-problem $\textbf{Arm}_i( {\bla})$ \eqref{eq:subprob} is computationally tractable and can be performed in parallel, since it does not involve the states and actions of other arms. In addition, the concatenation of each arm's optimal policy $\rho_i^*( {\bla})$ yields the optimal policy $\pi^*( {\bla})$. 


To solve the Dual problem \eqref{eq:saddle} for the optimal Lagrange multiplier $\bla^*$, one can iteratively update $\lambda_g$ via \eqref{eq:lambda} for all $g\in\sG$ \cite{nedic2008subgradient, zou2021minimizing}: 
\begin{align} 
     &\lambda_g^{(k+1)}=\Big[\lambda_g^{(k)} \nonumber  \\
    &\ +\delta_k \mathbb{E}_{\pi^*(\bla^{(k)})}\big[ \sum_{t=0}^\infty \beta^t\mathbb{I}(g_t=g)\big(\sum_{i=1}^Na_{i,t}-C_g\big)\big]\Big]^+,\label{eq:lambda}
\end{align}
where $k$ is the iteration number, 
$\delta_k$ is the step size, and $[x]^+\!:=\!\max\{x,0\}$. In \eqref{eq:lambda}, $\bla^{(k)}\!:=\! (\lambda_g^{(k)})_{g\in\sG}$, and $\pi^*({\bla}^{(k)})$ denotes the optimal policy under ${\bla}^{(k)}$, which can be obtained by solving the sub-problem $\textbf{Arm}_i(\bla^{(k)})$ \eqref{eq:subprob} for all arms $i\in[N]$. 
When $\bla^{(k)}$ converges, the corresponding optimal policy $\pi^*({\bla}^{(k)})$ solves the Relaxed problem \eqref{eq:relax}. 
The next subsection introduces the methods for solving 
the sub-problem $ \textbf{Arm}_i( {\bla})$ \eqref{eq:subprob} and  computing the expectation
in \eqref{eq:lambda}.  


\subsection{Solution of Sub-Problems and  Expectation Computation}

First, we consider solving the sub-problem $ \textbf{Arm}_i( {\bla})$ \eqref{eq:subprob}. 
For each arm $i\in[N]$ and a given Lagrange multiplier $ {\bla}$, let $V^*_{i, {\bla}}(g,s)$ be the optimal state-value function that equals $ \max_{\rho_i} \!\mathbb{E}_{\rho_i}\Big[\sum_{t=0}^\infty\beta^t(r_{i,t}- {\lambda}_{g_t}a_{i,t})|g_0 = g, s_{i,0}=s\Big]$. Then, the Bellman optimality equation is given by \eqref{eq:bellman}:
\begin{align} \label{eq:bellman}
 & V^*_{i, {\bla}}(g,s) = \max_{a\in\{0,1\}}\Big[ R_i(g,s,a)- {\lambda}_g a\nonumber\\
  &\quad  +\beta\!\sum_{s'\in\sS, g'\in\sG}G(g'|g)P_i(s'|g, s, a)V^*_{i, {\bla}}(g',s') \Big],
\end{align}
for all $g\in \sG,s\in\sS$. When the transition kernels $G(\cdot)$ and  $P_i(\cdot)$  are  {known},  one can compute $V^*_{i, {\bla}}(g,s)$ for all $g\in \sG, s\in\sS$ by solving the linear programming (LP) \eqref{eq:LP}: 
\begin{subequations} \label{eq:LP}
    \begin{align} 
   & (\textbf{LP}_i( {\bla})): \  \min_{(v_{g,s})_{g\in\sG,s\in\sS}}  \sum_{s\in\sS, g\in\sG}\,  v_{g,s}\\
 &  \quad \text{s.t. } \,  v_{g,s} \geq  R_i(g,s,a)- {\lambda}_g a \nonumber\\
    &\qquad\quad + \beta\!\sum_{ g'\in\sG,s'\in\sS}\! G(g'|g)P_i(s'|g, s, a)v_{g',s'},\nonumber \\
    &\qquad\qquad\qquad\forall g\in\sG, s\in\sS, a\in\sA,
\end{align}
\end{subequations}
where we let $v_{g,s} = V^*_{i, {\bla}}(g,s)$ for notational simplicity.

After obtaining the optimal state-value function $V^*_{i, {\bla}}(g,s)$ for all $g\in \sG, s\in\sS$, we can define the optimal $Q$-function $Q^*_{i, {\bla}}(g,s,a)$ by \eqref{eq:Qfunction}:
\begin{align} \label{eq:Qfunction}
    & Q^*_{i, {\bla}}(g,s,a) =  R_i(g,s,a)- {\lambda}_g a \nonumber\\
    &\qquad +\beta\!\sum_{ g'\in\sG,s'\in\sS}\! G(g'|g)P_i(s'|g, s, a)V^*_{i, {\bla}}(g',s'). 
\end{align}
Then, the optimal policy  $\rho_i^*( {\bla})$ that solves the sub-problem $\textbf{Arm}_i( {\bla})$ \eqref{eq:subprob} is deterministic and given by \eqref{eq:rho_policy}:
\begin{align} \label{eq:rho_policy}
\rho_i^*( {\bla})\!: a_{i,t} \!=\!   \begin{cases}
        1, &\!\! \text{if }  Q^*_{i, {\bla}}(g_t,\!s_{i,t},\!1) \!>\! Q^*_{i,{\bla}}(g_t,\!s_{i,t},\!0), \\
        0, &\!\! \text{otherwise}.
    \end{cases}
\end{align}

\vspace{2pt}
In terms of computing the expectation term in \eqref{eq:lambda},  it can be reformulated as \eqref{eq:gradient}: 
\begin{align}\label{eq:gradient}
   & \mathbb{E}_{\pi^*(\bla^{(k)})}\!\big[ \sum_{t=0}^\infty \beta^t\mathbb{I}(g_t\!=\!g)\big(\sum_{i=1}^Na_{i,t}-C_g\big)\big] \nonumber\\
   = & \sum_{i=1}^N \mathbb{E}_{\rho^*_i(\bla^{(k)})}\!\Big[ \sum_{t=0}^\infty\! \beta^t\mathbb{I}(g_t\!=\!g) a_{i,t}\Big] \!-\! C_g \mathbb{E}\big[\sum_{t=0}^\infty \beta^t\mathbb{I}(g_t\!=\!g)  \big] \nonumber \\
    = &  \sum_{i=1}^N A_{i,g}^{{(k)}}(g_0,s_0) - C_g B_g(g_0). 
\end{align}
Here, $A_{i,g}^{(k)}(\hat{g},\hat{s})\!:=\! \mathbb{E}_{\rho^*_i(\bla^{(k)})}\!\Big[ \sum_{t=0}^\infty\! \beta^t  \mathbb{I}(g_t\!=\!g) a_{i,t}|g_0\!=\!\hat{g},s_0\!=\!\hat{s}\Big]$ under the policy $\rho^*_i(\bla^{(k)})$ with the initial conditions of $(\hat{g},\hat{s})$ for   arm $i$ and global context $g$, and we define $B_g(\hat{g})\!:=\! \mathbb{E}\big[\sum_{t=0}^\infty \beta^t\mathbb{I}(g_t\!=\!g) |g_0=\hat{g}  \big]$. 

Moreover, the values $A_{i,g}^{(k)}(\hat{g},\hat{s})$ for all $g\in\sG,\hat{g}\in\sG, \hat{s}\in\sS$ can be obtained by solving a set of linear equations \eqref{eq:Ags}:
%
\begin{align} \label{eq:Ags}
   & A_{i,g}^{(k)}(\hat{g},\hat{s}) = \mathbb{I}(\hat{g}\!=\!g) a^*(\hat{g},\hat{s}) \nonumber \\
    &\ + \beta\!\!\sum_{ g'\in\sG,s'\in\sS}\!\! G(g'|\hat{g})P_i\big(s'|\hat{g},\hat{s},  a^*(\hat{g},\hat{s})\big)A_{i,g}^{(k)}(g',s'),
\end{align}
for all $ g\in\sG,\hat{g}\in\sG, \hat{s}\in\sS$.
Here, $a^*(\hat{g},\hat{s})$ is the action to take when the global context is $\hat{g}$ and the arm state is $\hat{s}$ 
under the policy $\rho^*_i(\bla^{(k)})$ \eqref{eq:rho_policy}. Similarly, the values $B_g(\hat{g})$ for all $g\in\sG,\hat{g}\in\sG$ can be obtained by solving a set of linear equations \eqref{eq:Bgg}:
\begin{align} \label{eq:Bgg}
   & B_{g}(\hat{g}) = \mathbb{I}(\hat{g}=g)    + \beta\sum_{ g'\in\sG} G(g'|\hat{g}) B_{g}(g'),
\end{align}
for all $g\in\sG,\hat{g}\in\sG$. Essentially, equations \eqref{eq:Ags} and \eqref{eq:Bgg} represent the  recursive relations for $A_{i,g}^{(k)}(\hat{g},\hat{s})$ and $B_{g}(\hat{g})$. 
Then, the values $A_{i,g}^{{(k)}}(g_0,s_0)$ and $ B_g(g_0)$ in \eqref{eq:gradient} can be directly obtained based on the initial condition of $(g_0,s_0)$.

\subsection{Index Policy for Primal Problem with Known  Models}

\begin{algorithm}[tb]
   \caption{Index Policy Algorithm for Solving the CRB Problem with Known Arm Models.}
   \label{alg:index_policy}
\begin{algorithmic}[1] 
\STATE \textbf{Initialization}:  $\bla^{(0)}\!:=\!(\lambda_g^{(0)})_{g\in\sG}\leftarrow \bm{0}$; $k\leftarrow 0$; $\epsilon>0$. 
\REPEAT \label{step:start}
\STATE Perform the following three steps for each arm $i\in[N]$ in parallel: 
 {\setlist{leftmargin=2mm}
\begin{itemize}
    \item [] 1) Solve $\textbf{LP}_i( {\bla}^{(k)})$ \eqref{eq:LP} to obtain $V^*_{i, {\bla}^{(k)}}(g,s)$;
    \item [] 2) Compute the $Q$-function $Q^*_{i, {\bla}^{(k)}}(g,s,a)$ by \eqref{eq:Qfunction}; 
    \item [] 3) Construct the optimal policy $\rho_i^*( {\bla}^{(k)})$ by \eqref{eq:rho_policy}.
\end{itemize}
}

\STATE Let $\pi^*(\bla^{(k)}):=(\rho_i^*( {\bla}^{(k)}))_{i\in[N]}$, and perform the update \eqref{eq:lambda} to obtain $\bla^{(k+1)}$, where the expectation is computed using \eqref{eq:gradient}-\eqref{eq:Bgg}. Let $k\leftarrow k+1$.

\UNTIL{the convergence $||\bla^{(k)} - \bla^{(k-1)}||\leq \epsilon$ is met. }

\STATE 
 Let $\bla^{*}\leftarrow \bla^{(k)}$.
 \label{step:end}

\FOR{time $t = 0, 1, 2, \hdots$} \label{step:deploy:start}
\STATE  Compute the index of each arm $i\in[N]$: 
\begin{align} \label{eq:index2}
     I_{i,t} \leftarrow  Q^*_{i,{\bla}^*}(g_t,s_{i,t},1)  - Q^*_{i,{\bla}^*}(g_t,s_{i,t},0).
\end{align}

\STATE Sort all arms such that $I_{(1),t}\geq I_{(2),t}\geq\cdots \geq I_{(N),t}$.
\STATE Activate the top $C_{g_t}$ arms $(1),(2),\cdots,(C_{g_t})$. 
\ENDFOR  \label{step:deploy:end}
\end{algorithmic}
\end{algorithm}

The proposed dual decomposition procedure above solves the Relaxed problem \eqref{eq:relax}. However, the actions generated by the optimal policy do not necessarily satisfy the per-time-step budget constraint \eqref{eq:primal:budget} of the Primal problem \eqref{eq:primal}. To obtain feasible actions with respect to the Primal problem \eqref{eq:primal}, we define $I_{i,t}$ as the index of each arm $i$ at time $t$ by \eqref{eq:index}:
\begin{align}\label{eq:index}
    I_{i,t}:=  Q^*_{i,{\bla}}(g_t,s_{i,t},1)  - Q^*_{i,{\bla}}(g_t,s_{i,t},0),\ \forall i\in[N], 
\end{align}
where $Q^*_{i,{\bla}}(\cdot)$ is the optimal $Q$-function given by \eqref{eq:Qfunction}. Essentially, the index $I_{i,t}$ indicates the net long-term expected reward of activating arm $i$ at time $t$. To meet the per-time-step budget constraint \eqref{eq:primal:budget} and maximize the total reward, our idea is to select the top $C_{g_t}$ arms with the highest index values to activate at each time $t$. Specifically, we propose an index policy 
(i.e., Algorithm \ref{alg:index_policy}) to solve the Primal problem \eqref{eq:primal}. In Algorithm \ref{alg:index_policy}, Steps \ref{step:start}-\ref{step:end} are devoted to finding the optimal Lagrange multiplier $\bla^*$, while Steps \ref{step:deploy:start}-\ref{step:deploy:end} implement the index policy to select arms at each time $t$.

In Algorithm \ref{alg:index_policy}, the index of each arm is calculated based on the solution of the Relaxed problem \eqref{eq:relax} rather than that of the original Primal problem \eqref{eq:primal}. Hence, we theoretically analyze the asymptotic optimality of the proposed index policy for solving the Primal problem \eqref{eq:primal} in Section \ref{sec:analysis}, and also demonstrate this key property via numerical simulations in Section \ref{sec:simu:asymp}. In addition, Algorithm \ref{alg:index_policy} is developed for the case when the arm transition kernel models are known, while the next sub-section further extends it to an online learning algorithm for the case with unknown arm models.

\subsection{Online CRB Learning with Unknown Models}

In the problem setting where arm transition kernels are unknown, building upon Algorithm \ref{alg:index_policy}, we propose a model-based online learning algorithm, Algorithm \ref{alg:learn}, to learn the arm models and make selection decisions. Specifically, Algorithm \ref{alg:learn} updates the arm transition kernel models after each epoch  (every $T$ times) in Step \ref{step:update}, when the empirical frequency is used to estimate the transition probability, and it 
 re-computes the optimal Lagrange multiplier $\bla^*_n$ in Step \ref{step:lambda}. In Steps \ref{step:deploy:start3}-\ref{step:deploy:end2}, 
Algorithm \ref{alg:learn} employs the same index policy as that in Algorithm \ref{alg:index_policy} for optimal arm selection,  when the $\epsilon$-greedy mechanism is used to balance exploration and exploitation. 

\begin{algorithm}[tb]
   \caption{Online Learning Algorithm for Solving the CRB Problem with Unknown Arm  Models.}
   \label{alg:learn}
\begin{algorithmic}[1] 
\STATE \textbf{Initialization}: Time window $T$; $\epsilon_n>0$; initial transition kernel $P_i^{0}(\cdot)$ of each arm $i\in[N]$.

\FOR{epoch $n = 0, 1, 2, \hdots$} \label{step:deploy:start2}

\STATE  Based on the up-to-date transition kernel model $P_i^{n}(\cdot)$ of each arm $i\in[N]$, follow Steps \ref{step:start}-\ref{step:end} in Algorithm \ref{alg:index_policy} to compute the optimal $\bla^{*}_n$. \label{step:lambda}
 
\FOR{time $t=nT, nT\!+\!1,\cdots, nT\!+\!T\!-\!1$} \label{step:deploy:start3}

\STATE With probability of $1-\epsilon_n$, 
{\setlist{leftmargin=3mm}
\begin{itemize}
    \item []- Compute the index $I_{i,t}$  \eqref{eq:index} of each arm $i\!\in\![N]$;
    \item [] - Sort all arms such that $I_{(1),t}\geq I_{(2),t}\geq\cdots \geq I_{(N),t}$ and activate the top $C_{g_t}$ arms. 
\end{itemize}
}
With probability of $\epsilon_n$, randomly activate $C_{g_t}$ arms. 

\ENDFOR \label{step:deploy:end2}

\STATE   For each arm $i\in[N]$, based on the observed state transitions, update its transition kernel model by:
\begin{align*}
    P^{n+1}_i(s_{i,t+1}\!=\!s'|g_t\!=\!g,s_{i,t}\!=\!s,a_{i,t}\!=\!a) \!=\! \frac{M_{s',g,s,a}^i}{M_{g,s,a}^i}, 
\end{align*}
where $M_{g,s,a}^i$ and $M_{s',g,s,a}^i$ are arm $i$'s cumulative historical counts of the context-state-action tuple $(g,s,a)$ and the state transition $(g,s,a)\to s'$. \label{step:update}

\ENDFOR  \label{step:deploy:end3}
\end{algorithmic}
\end{algorithm}

\section{Asymptotic Optimality Analysis} \label{sec:analysis}

In our proposed Algorithms \ref{alg:index_policy} and \ref{alg:learn}, the index policy is derived based on  the solution of the Relaxed problem \eqref{eq:relax} rather than that of the original Primal problem \eqref{eq:primal}. Hence, it is crucial to characterize the difference in solution between these two problems. Specifically, 
let $V_{\mathrm{Pri}}^N$ and $V_{\rel}^N$ be the optimal objective values of the Primal problem \eqref{eq:primal} and the Relaxed problem \eqref{eq:relax} with $N$ arms, respectively. Since the Relaxed problem \eqref{eq:relax} is a relaxation of the Primal problem \eqref{eq:primal}, we directly have $V_{\rel}^N\geq V_{\pri}^N$. Moreover, 
we will show that under certain assumptions, the per-arm reward difference $ ({V_{\rel}^N\!-\! V_{\pri}^N})/{N}$ asymptotically converges to zero as $N\!\to\! \infty$, namely \emph{asymptotic optimality} \cite{weber1990index,gast2020exponential}. 
For the analysis, 
we focus on the homogeneous CRB systems under periodic global contexts and make the following assumption. 
\begin{assumption}\label{ass:homo}
    Suppose that  the following assumptions hold for the contextual restless bandit (CRB) system:  
\begin{itemize}
    \item [(A1)] All arms share the same transition kernel and reward function, i.e.,
    $P_i(\cdot) \equiv P(\cdot)$ and $R_i(\cdot)\equiv R(\cdot)$ for all $i\in[N]$.
     \item [(A2)] The transition kernel $P(s'|g,s,a)$ is a rational number for all $s'\in\sS,g\in\sG, s\in\sS, a\in\sA$.
     \item [(A3)] The global context $g_t$ is periodic. That is, there exists an integer $T$ such that $g_{t+T}=g_t$ for all time $t$.  
     \item [(A4)] The budget $C_g$ scales linearly with the arm number $N$, i.e., $C_g = \alpha_gN $ with certain fixed $\alpha_g$,  for all $g\in\sG$. 
\end{itemize} 
\end{assumption}
These assumptions above are primarily made for theoretical analysis. The subsequent simulation results show that some assumptions, such as (A3), can be relaxed without compromising the asymptotic optimality.


Denote $\pi^*_{\rel}$ as the optimal policy for the Relaxed problem \eqref{eq:relax} with $N=1$. Due to assumptions (A1) and (A4), the optimal policy for the Relaxed problem \eqref{eq:relax} with $N\!>\!1$ is simply the application of $\pi^*_{\rel}$ to every arm independently. Thus, we have 
\begin{equation}
    V_{\rel}^N=N V_{\rel}^1 =  N\mathbb{E}_{\pi^*_{\rel}}\Big[\sum_{t=0}^\infty \beta^tr_{1,t}\Big].
\end{equation} 
Let $m^*_g$ be the steady-state distribution of $s_{i,t}$ under the global state $g_t=g$ given the policy $\pi^*_{\rel}$, namely,
\begin{align}
    m^*_g(s):=\lim_{t\rightarrow\infty}\mathbb{P}(s_{i,t}=s\Big|\,g_t=g, \pi^*_{\rel}), \ \forall s\in\sS.
\end{align}
Then, we establish the following lemma on the initial action.
\begin{lemma} \label{lemma:feasibleC_g}
 Given the initial global context $g_0 = g$ and suppose that the initial state $s_{i,0}$ of each arm $i\in [N]$ is chosen independently with the distribution $\mathbb{P}(s_{i,0}=s) = m^*_g(s)$, then, under the policy $\pi^*_{\rel}$, we have 
    \begin{equation}
        \mathbb{E}_{\pi^*_{\rel}}\Big[\sum_{i=1}^N a_{i,0}\Big]\leq C_g.
    \end{equation} 
\end{lemma}
\noindent
\textit{Proof}. 
    By the definition of $m^*_g$, we have $ \mathbb{P}(s_{i,t}=s)=m^*_g(s)$ for all $t>0$ when $g_t=g$. Hence,  
  $\mathbb{E}_{\pi^*_{\rel}}\big[\sum_{i=1}^N a_{i,t}\big]=\mathbb{E}_{\pi^*_{\rel}}\big[\sum_{i=1}^N a_{i,0}\big]$ for all $t>0$ when $g_t=g$. Then, 
    \begin{align}
      \mathbb{E}\Big[\sum_{t=0}^\infty \beta^t\mathbb{I}(g_t\!=\!g)\Big]&C_g \geq    \mathbb{E}_{\pi^*_{\rel}}\Big[\sum_{t=0}^\infty \beta^t\mathbb{I}(g_t\!=\!g)\big(\sum_{i=1}^Na_{i,t}\big)\Big] \nonumber \\
        =\, &\mathbb{E}\Big[\sum_{t=0}^\infty \beta^t\mathbb{I}(g_t=g)\Big]\mathbb{E}_{\pi^*_{\rel}}\Big[\sum_{i=1}^Na_{i,t}\Big], \nonumber 
        \\
        =\, &\mathbb{E}\Big[\sum_{t=0}^\infty \beta^t\mathbb{I}(g_t=g)\Big]\mathbb{E}_{\pi^*_{\rel}}\Big[\sum_{i=1}^Na_{i,0}\Big], \nonumber
    \end{align}
where the first inequality is due to the constraint \eqref{eq:relax:budget}. Thus, we have $ \mathbb{E}_{\pi^*_{\rel}}\Big[\sum_{i=1}^N a_{i,0}\Big]\leq C_g.$
\qed

\vspace{2pt}

Next, we establish the following theorem that shows the asymptotical optimality between the solutions of the Primal problem \eqref{eq:primal} and the Relaxed problem \eqref{eq:relax}.
\begin{theorem} \label{thm:optimality}
 Suppose that the initial global context $g_{0}$ is chosen uniformly at random from $\sG$ and the initial state $s_{i,0}$ of each arm $i\in[N]$ is chosen independently with the distribution $\mathbb{P}(s_{i,0}\!=\!s|g_0\!=\!g) = m^*_g(s)$, then, under Assumption \ref{ass:homo}, we have 
    \begin{equation}
        V_{\rel}^N\geq V_{\pri}^N\geq V_{\rel}^N - \mathcal{O}(\sqrt{N}).
    \end{equation}
\end{theorem}
The proof of Theorem \ref{thm:optimality} is provided in Appendix \ref{app:proof:thm1}. Theorem \ref{thm:optimality} implies that as the number of arms $N$ increases, the optimal per-arm rewards of the Primal problem \eqref{eq:primal} and the Relaxed problem \eqref{eq:relax}, namely $V_{\rel}^N/N$ and  $V_{\pri}^N/N$, 
asymptotically become the same. Furthermore, let $ V^N_{\ind} \!:=\! \mathbb{E}_{\pi_\ind}\!\Big[\!\sum_{t=0}^\infty\! \sum_{i=1}^N\! \beta^t r_{i,t}\Big]$ be the expected cumulative reward using the index policy (Algorithm \ref{alg:index_policy}).  Since the index policy generates feasible solutions for the Primal problem \eqref{eq:primal}, we have $V_{\rel}^N \geq V_{\pri}^N\geq V_{\ind}^N.$ 
The simulations in Section \ref{sec:simu:asymp} demonstrate 
the asymptotic optimality of the index policy.



\section{Numerical Simulations}\label{sec:simulation}

This section takes the DR decision-making problem, introduced in Section \ref{sec:drapp}, as a practical application to demonstrate the performance of our proposed CRB framework and index policy algorithm via numerical simulations.



In the DR simulations, we consider a discrete global context 
with $g\in\sG\! = \!\{1,2,\cdots,6\}$ to reflect different levels of ambient temperature and electricity prices,  
and a uniform transition distribution
$G(g'|g)\!:=\! 1/|\sG|$. We set the number of users as $N \!=\! 500$ and the selection ratio as $\alpha_g \!=\! 0.2$, indicating that at most $\alpha_gN$ users can be selected at each time $t$.
For the state $s_{i,t}\!:=\!( z_{i,t}, x_{i,t})$ of each user $i$ defined in \eqref{eq:state}, let $x_{i,t}\in\{1,2,\cdots,4\}$ be the level of user DR fatigue. 
The reward function of each user  $i$ is designed as  $R_i(\cdot)\! =\! \frac{a_{i,t} z_{i,t}l_i}{({g_{t}- x_{i,t}})^2+1} $,\footnote{This reward function is designed to reflect the influence of global context and fatigue effect on users' load reduction, while other reasonable reward functions can also be used for simulation. In practice, such a reward function is unknown \emph{a priori} and needs to be learned based on real observations. } and $l_i$ is randomly sampled from the uniform distribution $\text{Unif}([8, 12])$ (kWh) to capture the diversity in flexible load availability across users. 
Let the discount factor be $\beta \!=\!0.97$, and we use a finite time horizon $T = 300$ to approximate the infinite time horizon in simulations. 


\subsection{Convergence of Dual Decomposition}

\begin{figure}
    \centering
    \includegraphics[scale=0.37]{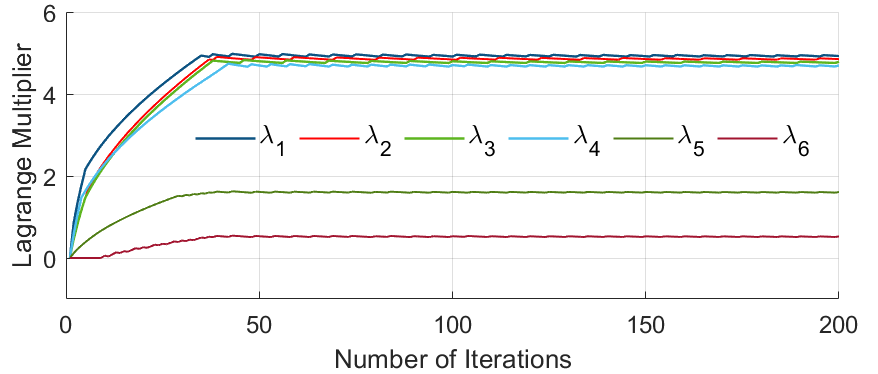}
    \caption{Convergence of the Lagrange multiplier $\bla\!:=\!(\lambda_g)_{g\in\sG}$ with the dual decomposition method.}
    \label{fig:lambda}
\end{figure}

We first study the convergence of the dual decomposition method that is used for solving the optimal Lagrange multiplier $\bla^*$. It corresponds to Steps \ref{step:start}-\ref{step:end} in Algorithm \ref{alg:index_policy}. Figure \ref{fig:lambda} illustrates the convergence results of $\bla\!:=\!\{\lambda_1,\cdots,\lambda_6\}$, where each component corresponds to a global context $g\in\sG$. It is observed that all Lagrange multipliers converge rapidly within tens of iterations. Besides, the optimal values of $\lambda_g^*$ vary across different global contexts, which correspond to the expected reward that can be obtained under the global context $g$. It is consistent with the intuition that the optimal Lagrange multiplier $\lambda_g^*$ can be interpreted as the activation cost that the decision-maker is willing to pay for pulling an arm under the global context $g$ \cite{whittle1988restless}.


\subsection{Asymptotic Optimality of Index Policy}\label{sec:simu:asymp}

\begin{figure}
    \centering
    \includegraphics[scale=0.36]{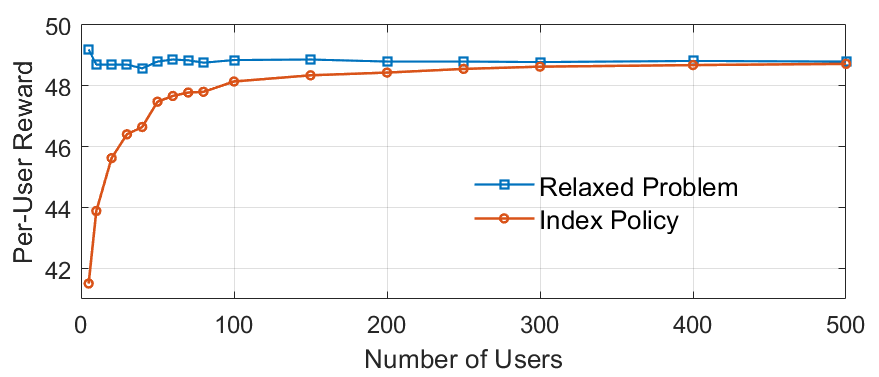}
    \caption{Comparison between the per-user reward $V^N_{\rel}/N$ and $V^N_{\mathrm{Ind}}/N$ of the Relaxed problem \eqref{eq:relax} and of the index policy (Algorithm \ref{alg:index_policy}).}
    \label{fig:asmp}
\end{figure}

This part demonstrates the asymptotic optimality of our proposed index policy algorithm (Algorithm \ref{alg:index_policy}). We gradually increase the number of users $N$ from 5 to 500, while keeping the same selection ratio $\alpha_g =0.2$. In each case with different $N$, we compute the corresponding optimal objective value $V_{\rel}^N$ of the Relaxed problem \eqref{eq:relax}; then we run 500 rounds of simulations and take the average of the cumulative rewards to estimate the  
expected cumulative reward $V^N_{\ind}$ using the index policy.
The results of per-user reward $V_{\rel}^N/N$ and $V^N_{\ind}/N$ of the Relaxed problem \eqref{eq:relax} and the index policy are illustrated in 
Figure \ref{fig:asmp}. It is seen that the per-user reward $V^N_{\ind}/N$ of the index policy becomes larger as the user number $N$ increases, and asymptotically converges to the per-user reward $V_{\rel}^N/N$ of the Relaxed problem \eqref{eq:relax}. Since  $V_{\rel}^N \geq V_{\pri}^N\geq V_{\ind}^N$ by definitions, the simulation results also imply that $V_{\ind}^N/N \to V_{\pri}^N/N$ as $N\to \infty$, namely the asymptotical optimality of the index policy for solving the Primal problem \eqref{eq:primal}. Additionally, in Figure \ref{fig:asmp}, the small variations in the per-user reward of the Relaxed problem are caused by the random sampling of the initial conditions.

\subsection{Performance Comparison with Restless Bandits}

\begin{figure}
    \centering
    \includegraphics[scale=0.36]{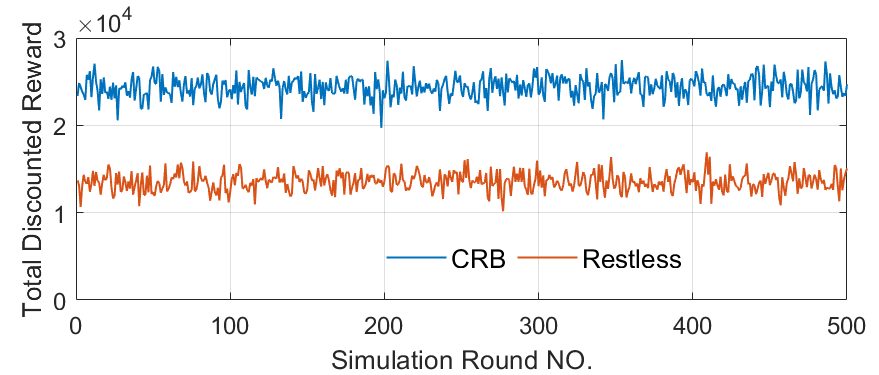}
    \caption{Comparison of the total discounted rewards between the CRB method and the traditional restless bandits method.}
    \label{fig:comp}
\end{figure}

We compare the proposed CRB approach with the traditional restless bandits method, which does not incorporate contextual information. For the restless bandits method, we let $P^{\mathrm{rb}}_i(s'|s,a) \!:=\! \mathbb{E}_{g\sim h(g)}[P_i(s'|g,s,a)] $ and $R^{\mathrm{rb}}_i(s,a)\!:=\!\mathbb{E}_{g\sim h(g)}[R_i(g,s,a)]$ be the transition kernel and reward function for arm $i\in[N]$, where $h(g)$ is the stationary distribution of the global context. Then, we derive a similar index policy, yet without consideration of global context, for the restless bandits, and compare it with Algorithm \ref{alg:index_policy} through simulation tests. Figure \ref{fig:comp} illustrates the comparison of the total discounted reward, i.e., $\sum_{t=1}^T\sum_{i=1}^N \beta^t r_{i,t} $, between the CRB-based index policy and the restless bandits-based index policy over 500 rounds of simulations. 
It is seen that the CRB approach achieves much higher rewards than the restless bandits method in all rounds of simulations, due to the incorporation of global contexts. 
The average total discounted reward over 500 rounds using the CRB approach amounts to $2.44\times 10^4$ kWh of load reduction, significantly surpassing that achieved by the restless bandits method, which stands at $1.36\times 10^4$ kWh of load reduction.

\section{Conclusion} \label{sec:conclusion}

In this paper, driven by the practical needs of demand response applications, we introduce a novel Contextual Restless Bandits (CRB) framework that models both the internal state transitions of each arm and the influence of external global environmental contexts. We then develop a scalable index policy algorithm for optimal arm selection decisions in the CRB problem. The proposed CRB framework and the index policy algorithm are applied to the demand response decision-making problem. 
Simulation results show that the dual decomposition method used for generating the index policy achieves rapid convergence, and the proposed index policy algorithm exhibits the property of asymptotic optimality. 
Furthermore, the proposed CRB approach consistently outperforms traditional restless bandits methods in terms of rewards, which underscores the importance of integrating global contextual information.



\appendices
\section{Proof of Theorem \ref{thm:optimality}} \label{app:proof:thm1}

\noindent
\textit{Proof}. 
Since every feasible policy for the Primal problem \eqref{eq:primal} is a feasible policy for the Relaxed problem \eqref{eq:relax}, we directly have $V_{\rel}^N\geq V_{\pri}^N$. 
For a given $N$, let $M_t^N(s)$ be the number of arms with state $s_{i,t}=s$ at time $t$, and define $M_0^N(\sS)\!:=\!(M_0^N(s))_{s\in\sS}$ to capture the number of arms at each state $s$ initially. 
Let
 $\tilde{V}^N_{\pri}(M_0^N(\sS), g_0)$ and $\tilde{V}^N_{\rel}(M_0^N(\sS), g_0)$ be the values of $V^N_{\pri}$ and $V^N_{\rel}$ under the initial condition of $g_0$ and $M_0^N(\sS)$, respectively. 
Thus, by definition, 
\begin{align*}
    V^N_{\pri}\!=\! 
 \mathbb{E}\Big[\tilde{V}^N_{\pri}\big(M_0^N(\sS), g_0\big)\Big],
 V^N_{\rel}\!=\! 
 \mathbb{E}\Big[\tilde{V}^N_{\rel}\big(M_0^N(\sS), g_0\big)\Big],
\end{align*}
where $g_0$ is chosen uniformly at random from $\sG$ and $M_0^N(\sS)$ is resulted by 
randomly and independently sampling the initial state of each arm using
the distribution $m^*_g(s)$. 

 Our proof follows the proof of \cite[Theorem 2.1]{gast2020exponential} and encompasses the following three steps.

    \textit{Step 1):} Due to (A2) in Assumption \ref{ass:homo}, $m^*_g(s)$ is a rational number for all $g\in\sG,s\in\sS$. Thus, one can pick a sufficiently large $N$ such that $Nm^*_g(s)$ is an integer for all $g\in\sG,s\in\sS$.
  For a given $g$, we construct a system such that $g_0=g$ and $M_0^N(s)=Nm^*_g(s)$. Since there are precisely $Nm^*_g(s)$ arms with $s_{i,0}=s$, by Lemma~\ref{lemma:feasibleC_g}, applying $\pi^*_{\rel}$ to each arm $i\in[N]$ will activate at most $C_g$ arms and thus satisfies the Primal budget constraint \eqref{eq:primal:budget}. 
  Then, we consider a new policy for the Primal problem \eqref{eq:primal}:  
 at time $t=0$, the policy applies $\pi^*_{\rel}$ to each arm; starting from time $t=1$, the policy follows the optimal policy $\pi_{\pri}^*$ for solving the Primal problem \eqref{eq:primal}. 
 This policy is clearly feasible for the Primal problem but it may be a sub-optimal policy. Thus, we have
  \begin{align}
        &\tilde{V}^N_{\pri}\big(  \big(Nm^*_{g}(s)\big)_{s\in\sS}, g\big) 
        \geq N\mathbb{E}_{\pi^*_{\rel}}\big[r_{1,0}\big|g_0=g, s_{1,0}\!\sim\! m^*_{g}\big]\nonumber\\
        &\qquad  + \beta \mathbb{E}\Big[\tilde{V}^N_{\pri}\big(M^{N}_1(\sS), g_1\big)\big|g_0=g\Big], \quad \forall g\in\sG.\label{equation:optimality:eq1}
    \end{align}
    
Note that $\mathbb{E}_g\Big[\mathbb{E}_{\pi^*_{\rel}}\big[r_{1,0}\big|g_0\!=\!g, s_{1,0}\!\sim\! m^*_{g}\big]\Big]$ is the steady-state expected per-step reward of applying $\pi^*_{\rel}$ to one arm. Thus, we have 
    \begin{align}
V^N_{\rel}&=N\mathbb{E}_{\pi^*_{\rel}}\Big[\sum_{t=0}^\infty \beta^tr_{1,t}\Big]
      \nonumber  \\
     &=   \frac{N}{1-\beta}\mathbb{E}_g\Big[\mathbb{E}_{\pi^*_{\rel}}\big[r_{1,0}\big|g_0\!=\!g, s_{1,0}\!\sim\! m^*_{g}\big]\Big].\label{equation:optimality:eq2}
    \end{align}
    Taking the expectation over $g$ on both sides of  (\ref{equation:optimality:eq1}) and incorporating (\ref{equation:optimality:eq2}), we obtain
    \begin{align}
        &V^N_{\pri} 
      =  \mathbb{E}_g\Big[\tilde{V}^N_{\pri}\big(  \big(Nm^*_{g}(s)\big)_{s\in\sS}, g\big) \Big] \nonumber\\
       \geq  & N\mathbb{E}_g\Big[\mathbb{E}_{\pi^*_{\rel}}\big[r_{1,0}\big|g_0\!=\!g, s_{1,0}\!\sim\! m^*_{g}\big]\Big]\nonumber \\
       &\qquad \qquad +\beta \mathbb{E}_g\Big[\mathbb{E}\big[\tilde{V}^N_{\pri}(M^{N}_1(\sS), g_1)\big|g_0=g\big]\Big] \nonumber \\
        = &(1\!-\!\beta)V^N_{\rel}+ \beta \mathbb{E}_g\Big[\mathbb{E}\big[\tilde{V}^N_{\pri}(M^{N}_1(\sS), g_1)\big|g_0=g\big]\Big].\label{equation:optimality:eq3}
    \end{align}
Since $V^N_{\pri}= (1\!-\!\beta) V^N_{\pri} \!+\! \beta \mathbb{E}_g\Big[\tilde{V}^N_{\pri}\big(  \big(Nm^*_{g}(s)\big)_{s\in\sS}, g\big) \Big]$, 
\begin{align}\label{eq:eq1}
    V^N_{\pri} - V^N_{\rel} &\geq \frac{\beta}{1-\beta}\mathbb{E}_g\Big[\mathbb{E}\big[\tilde{V}^N_{\pri}(M^{N}_1(\sS), g_1)\big|g_0=g\big] \nonumber\\ 
   & \qquad\quad- \tilde{V}^N_{\pri}\big(  \big(Nm^*_{g}(s)\big)_{s\in\sS}, g\big)\Big].  
\end{align}
 Since $g_0$ is chosen uniformly at random from $\sG$ and $g_t$ is periodic due to Assumption \ref{ass:homo} (A3), $g_1$ is also uniformly distributed from $\mathcal{G}$. Therefore, by \eqref{eq:eq1}, we have 
    \begin{align}
        V^N_{\pri}-V^N_{\rel}
        \geq &\frac{\beta}{1-\beta}\mathbb{E}_g\Big[\mathbb{E}\big[\tilde{V}^N_{\pri}\big(M^{N}_1(\sS), g_1\big)\nonumber\\
        &-\tilde{V}^N_{\pri}\big((Nm^*_{g_1}(s))_{s\in\sS}, g_1)\big)\big|g_0=g\big]\Big].\label{equation:optimality:eq4}
    \end{align}
Then, we will further bound the term $\tilde{V}^N_{\pri}\big(M^{N}_1(\sS), g_1\big)-\tilde{V}^N_{\pri}\big((Nm^*_{g_1}(s))_{s\in\sS}, g_1)$ in Step 2) and Step 3) below.

 \vspace{2pt}
\textit{Step 2):} This step establishes the claim that there exists some constant $\delta>0$ such that, 
    \begin{align}
        &\tilde{V}^N_{\pri}\big( M^{N}_1(\sS), g_1\big)-\tilde{V}^N_{\pri}\big((Nm^*_{g_1}(s))_{s\in\sS}, g_1)\big)\nonumber\\
        \leq\, & \delta\sum_{s\in\sS} \big|M^{N}_1(s)-Nm^*_{g_1}(s)\big|.\label{equation:optimality:step2}
    \end{align}
The proof of this claim directly follows the proof of \cite[Theorem 2.1]{gast2020exponential} and thus is omitted here.

    \vspace{2pt}

    \textit{Step 3):} This step establishes the claim that for all $g\in\sG$,
    \begin{align}
        \mathbb{E}\Big[\sum_{s\in\sS} \big|M^{N}_1(s)-Nm^*_{g_1}(s)\big|\Big|g_0=g\Big]=\mathcal{O}(\sqrt{N}). \label{equation:optimality:eq5}
    \end{align}
    To prove equation (\ref{equation:optimality:eq5}), we note that ${M^{N}_1(s)}/{N}$ is the empirical distribution of the states of arms when we apply the policy $\pi^*_{\rel}$ to the steady-state distribution $m^*_{g_0}$. Since $g_t$ is periodic, applying $\pi^*_{\rel}$ to the steady-state distribution $m^*_{g_0}$ yields the steady-state distribution  $m^*_{g_0}$. Hence, $\mathbb{E}\big[M^{N}_1(s)|g_0=g]=Nm^*_{g_1}(s)$. Moreover, since the state of each arm evolves independently, the variance of $M^{N}_1(s)$, i.e., $\mathbb{E}\big[\sum_{s\in\sS} (M^{N}_1(s)-Nm^*_{g_1}(s))^2\big|g_0=g\big]$ is on the order of $\mathcal{O}(N)$. Therefore, equation (\ref{equation:optimality:eq5}) holds.

    Combining (\ref{equation:optimality:eq4}), (\ref{equation:optimality:step2}), and (\ref{equation:optimality:eq5}) completes the proof. \qed

\bibliography{ref}

\begin{thebibliography}{10}
\providecommand{\url}[1]{#1}
\csname url@samestyle\endcsname
\providecommand{\newblock}{\relax}
\providecommand{\bibinfo}[2]{#2}
\providecommand{\BIBentrySTDinterwordspacing}{\spaceskip=0pt\relax}
\providecommand{\BIBentryALTinterwordstretchfactor}{4}
\providecommand{\BIBentryALTinterwordspacing}{\spaceskip=\fontdimen2\font plus
\BIBentryALTinterwordstretchfactor\fontdimen3\font minus \fontdimen4\font\relax}
\providecommand{\BIBforeignlanguage}[2]{{%
\expandafter\ifx\csname l@#1\endcsname\relax
\typeout{** WARNING: IEEEtran.bst: No hyphenation pattern has been}%
\typeout{** loaded for the language `#1'. Using the pattern for}%
\typeout{** the default language instead.}%
\else
\language=\csname l@#1\endcsname
\fi
#2}}
\providecommand{\BIBdecl}{\relax}
\BIBdecl

\bibitem{slivkins2019introduction}
A.~Slivkins \emph{et~al.}, ``Introduction to multi-armed bandits,'' \emph{Foundations and Trends in Machine Learning}, vol.~12, no. 1-2, pp. 1--286, 2019.

\bibitem{bouneffouf2019survey}
D.~Bouneffouf and I.~Rish, ``A survey on practical applications of multi-armed and contextual bandits,'' \emph{arXiv preprint arXiv:1904.10040}, 2019.

\bibitem{misra2019dynamic}
K.~Misra, E.~M. Schwartz, and J.~Abernethy, ``Dynamic online pricing with incomplete information using multiarmed bandit experiments,'' \emph{Marketing Science}, vol.~38, no.~2, pp. 226--252, 2019.

\bibitem{chen2020online}
X.~Chen, Y.~Nie, and N.~Li, ``Online residential demand response via contextual multi-armed bandits,'' \emph{IEEE Control Systems Letters}, vol.~5, no.~2, pp. 433--438, 2020.

\bibitem{li2020reliability}
Y.~Li, Q.~Hu, and N.~Li, ``A reliability-aware multi-armed bandit approach to learn and select users in demand response,'' \emph{Automatica}, vol. 119, p. 109015, 2020.

\bibitem{villar2015multi}
S.~S. Villar, J.~Bowden, and J.~Wason, ``Multi-armed bandit models for the optimal design of clinical trials: benefits and challenges,'' \emph{Statistical science: A Review Journal of the Institute of Mathematical Statistics}, vol.~30, no.~2, p. 199, 2015.

\bibitem{elena2021survey}
G.~Elena, K.~Milos, and I.~Eugene, ``Survey of multiarmed bandit algorithms applied to recommendation systems,'' \emph{International Journal of Open Information Technologies}, vol.~9, no.~4, pp. 12--27, 2021.

\bibitem{whittle1988restless}
P.~Whittle, ``Restless bandits: Activity allocation in a changing world,'' \emph{Journal of Applied Probability}, vol.~25, no.~A, pp. 287--298, 1988.

\bibitem{weber1990index}
R.~R. Weber and G.~Weiss, ``On an index policy for restless bandits,'' \emph{Journal of Applied Probability}, vol.~27, no.~3, pp. 637--648, 1990.

\bibitem{zhou2015survey}
L.~Zhou, ``A survey on contextual multi-armed bandits,'' \emph{arXiv preprint arXiv:1508.03326}, 2015.

\bibitem{chen2022reinforcement}
X.~Chen, G.~Qu, Y.~Tang, S.~Low, and N.~Li, ``Reinforcement learning for selective key applications in power systems: Recent advances and future challenges,'' \emph{IEEE Transactions on Smart Grid}, vol.~13, no.~4, pp. 2935--2958, 2022.

\bibitem{qdr2006benefits}
{U.S. Department of Energy}, ``Benefits of demand response in electricity markets and recommendations for achieving them,'' \emph{Washington, DC, USA, Tech. Rep}, vol. 2006, p.~95, 2006.

\bibitem{xu2018promoting}
X.~Xu, C.-f. Chen, X.~Zhu, and Q.~Hu, ``Promoting acceptance of direct load control programs in the united states: Financial incentive versus control option,'' \emph{Energy}, vol. 147, pp. 1278--1287, 2018.

\bibitem{fell2015public}
M.~J. Fell, D.~Shipworth, G.~M. Huebner, and C.~A. Elwell, ``Public acceptability of domestic demand-side response in great britain: The role of automation and direct load control,'' \emph{Energy research \& social science}, vol.~9, pp. 72--84, 2015.

\bibitem{shi2019estimating}
Q.~Shi, C.-F. Chen, A.~Mammoli, and F.~Li, ``Estimating the profile of incentive-based demand response (ibdr) by integrating technical models and social-behavioral factors,'' \emph{IEEE Transactions on Smart Grid}, vol.~11, no.~1, pp. 171--183, 2019.

\bibitem{kim2011common}
J.-H. Kim and A.~Shcherbakova, ``Common failures of demand response,'' \emph{Energy}, vol.~36, no.~2, pp. 873--880, 2011.

\bibitem{taylor2013index}
J.~A. Taylor and J.~L. Mathieu, ``Index policies for demand response,'' \emph{IEEE Transactions on Power Systems}, vol.~29, no.~3, pp. 1287--1295, 2013.

\bibitem{chen2021online}
X.~Chen, Y.~Li, J.~Shimada, and N.~Li, ``Online learning and distributed control for residential demand response,'' \emph{IEEE Transactions on Smart Grid}, vol.~12, no.~6, pp. 4843--4853, 2021.

\bibitem{yu2018deadline}
Z.~Yu, Y.~Xu, and L.~Tong, ``Deadline scheduling as restless bandits,'' \emph{IEEE Transactions on Automatic Control}, vol.~63, no.~8, pp. 2343--2358, 2018.

\bibitem{yu2016demand}
Z.~Yu and L.~Tong, ``Demand response via large scale charging of electric vehicles,'' in \emph{2016 IEEE Power and Energy Society General Meeting (PESGM)}.\hskip 1em plus 0.5em minus 0.4em\relax IEEE, 2016, pp. 1--5.

\bibitem{liang2024bayesian}
B.~Liang, L.~Xu, A.~Taneja, M.~Tambe, and L.~Janson, ``A bayesian approach to online learning for contextual restless bandits with applications to public health,'' \emph{arXiv preprint arXiv:2402.04933}, 2024.

\bibitem{adelman2008relaxations}
D.~Adelman and A.~J. Mersereau, ``Relaxations of weakly coupled stochastic dynamic programs,'' \emph{Operations Research}, vol.~56, no.~3, pp. 712--727, 2008.

\bibitem{nedic2008subgradient}
A.~Nedic and A.~Ozdaglar, ``Subgradient methods in network resource allocation: Rate analysis,'' in \emph{2008 42nd Annual Conference on Information Sciences and Systems}.\hskip 1em plus 0.5em minus 0.4em\relax IEEE, 2008, pp. 1189--1194.

\bibitem{zou2021minimizing}
Y.~Zou, K.~T. Kim, X.~Lin, and M.~Chiang, ``Minimizing age-of-information in heterogeneous multi-channel systems: A new partial-index approach,'' in \emph{Proceedings of the twenty-second international symposium on theory, algorithmic foundations, and protocol design for mobile networks and mobile computing}, 2021, pp. 11--20.

\bibitem{gast2020exponential}
N.~Gast, B.~Gaujal, and C.~Yan, ``Exponential convergence rate for the asymptotic optimality of whittle index policy,'' \emph{arXiv preprint arXiv:2012.09064}, 2020.

\end{thebibliography}





\end{document}